# Text Classification with Lexicon from Pre-Attention Mechanism

QINGBIAO LI1, CHUNHUA WU1, KANGFENG ZHENG1
1School of Cyberspace Security, Beijing University of Posts and Telecommunications, Beijing 100876, China
.

**ABSTRACT**

Text classification is an important task in the field of natural language processing (NLP). A comprehensive and high-quality lexicon plays a crucial role in traditional text classification approaches. And it improves the utilization of the linguistic knowledge. Although it is helpful for the task, the lexicon has got little attention in recent neural network models. Firstly, getting a high-quality lexicon is not easy. We lack an effective automated lexicon extraction method, and most lexicons are hand crafted, which is very inefficient for big data. What's more, there is no an effective way to use a lexicon in a neural network. To address those limitations, we propose a Pre-Attention mechanism for text classification in this paper, which can learn attention of different words according to their effects in the classification tasks. The words with different attention can form a domain lexicon. Experiments on three benchmark text classification tasks show that our models get competitive result comparing with the state-of-the-art methods. We get 90.5% accuracy on Stanford Large Movie Review dataset, 82.3% on Subjectivity dataset, 93.7% on Movie Reviews. And compared with the text classification model without Pre-Attention mechanism, those with Pre-Attention mechanism improve by 0.9%-2.4% accuracy, which proves the validity of the Pre-Attention mechanism. In addition, the Pre-Attention mechanism performs well followed by different types of neural networks (e.g., convolutional neural networks and Long Short-Term Memory networks). For the same dataset, when we use Pre-Attention mechanism to get attention value followed by different neural networks, those words with high attention values have a high degree of coincidence, which proves the versatility and portability of the Pre-Attention mechanism. we can get stable lexicons by attention values, which is an inspiring method of information extraction.

## I. INTRODUCTION

Text classification is an important task in the field of NLP, which plays an important role in many practical applications, such as email categorization, web search, file classification, etc [1,2]. A lot of researches have been done in text classification. SynTime1 [3] model uses three main syntactic token types to recognize time expression. Lifelong learning model [4] for sentiment classification adopts a Bayesian optimization framework. In the traditional text classification method, a popular and common method of expressing text is the bag-of-words. However, the bag-of-words method loses the order of the words and ignores the semantics of the words. The n-gram model is very popular in statistical language models and usually performs well [5]. However, the n-gram model has a large defect that is affected by data sparsity [6].

Recently, neural network methods are becoming more and more popular, for it can train a more complex model on a large dataset. And it can also overcome the data sparsity problem of the n-gram model [6]. Neural network models based on deep learning have achieved significant success on many NLP tasks, including learning distributed word, sentence and document representation[7], parsing[8], statistical machine translation[9], sentiment classification[10], etc. In the field of text classification, neural networks have been widely used and perform well. Some fast and efficient neural network-based methods have been proposed. For example, fast text is a linear word-level model with a rank constraint and fast loss approximation, which achieved competitive results with a simple structure. [11].

Though deep neural networks have gained great success in text classification field, these methods do not make full use of the linguistic knowledge, because not all words have the same importance in text classification. For example, a high-quality sentiment lexicon is very important for a sentiment classification task, and it would be easier for us to classify one's texts from other people's texts if we have the lexicon of his most-used words. The traditional text classification approaches used the classification lexicon including those words playing a crucial role for text classification task, [12] which has has a positive impact on improving classification accuracy. For example, emotion features extracted using the knowledge of the general purpose emotion lexicons (GPELs), when combined with traditional bag-of-words features improved emotion classification significantly [13], [14]. But the lexicon for classification has received little attention in recent neural network models. There are two main difficulties. Firstly, a high-quality lexicon is hard to obtain. In other words, it is difficult to find those words that are important for a classification task. We lack an effective automated lexicon extraction method, and most lexicons are hand crafted. For example, existing GPELS such as WordNet-Affect (WNA) [15], EmoSenticNet (ESN) [16] and NRC word-emotion lexicon [17], which are hand crafted, associate between words and emotions identified by Ekman and Plutchik. However, the efficiency of manual extraction is very low, especially when the amount of data is large. Secondly, there is no an effective way to use a lexicon in a neural network. Some methods using neural networks try to solve those problems with a attention mechanism[18, 19, 20]. But most traditional attention mechanisms rely on the RNN or encoding-decoding structures. Because they apply the attention mechanisms on the output states of RNN, it is difficult to migrate the same attention structure between neural networks of different structure, which limits the use of attention mechanisms. In addition, as getting the attention of a word, the attention mechanism has considered the context of the word, which leads different attention values for ths same word in different sentences. So, for a classification task on a database, we can not get which word is more important for the task by comparing the attention values of two words, and we can not get a lexicon, either.

To address the aforementioned limitations, we propose a Pre-Attention mechanism, which can automatically find a lexicon for a classification task and integrate it into the neural networks. The mechanism is located between the word vectors and the classification neural network. Firstly, we get the text representation with the Pre-Attention mechanism. Then we push the text representation to CNN and LSTM for classification. In addition, for each word in a database, we can calculate the accordingly attention value, which reflects the contribution of the word to the classification task, and we can get a lexicon according it. Compared with words with low attention values, those words with high attention values are more likely to belong to the lexicon. Our contributions are summarized below:

1) We provide a way to integrate a lexicon into deep neural networks for classification. The Pre-Attention mechanism can automatically pay different attention values to words according to their importance for a classification task, which improves the utilization of linguistic knowledge.
2) We provide a method of getting a stable lexicon through neural networks. For a text classification task, after getting the words' attention values by Pre-Attention mechanism, we can get a lexicon for classification. Experiments showed that even with different post-classification methods the lexicon is stable.

The rest of the paper is organized as follows. Section II presents related work. Section III describes model architecture. Section IV outlines the experimental setup. Section V discusses the empirical results and analysis. Finally, section VI presents the conclusion and future work.

## II. RELATED WORK

Natural Language Processing (NLP) is a sub-area of artificial intelligence (AI), which is also one of the most difficult problem in AI. With the development of the Internet, the amount of text data in the network has increased rapidly. NLP is becoming more and more important. Text classification is significant for NLP systems, where there have been an enormous amount of researches.

A simple and efficient baseline method for text classification is to train a linear classifier (e.g., a Support Vector Machines and logistic regression) after represent the sentence as a bag-of-words. However, the bag-of-words method loses the order of the words and ignores the semantics of the words. N-gram model is another popular method to represent a sentence, which usually performs well [21]. In addition, there are some other popular methods to obtain better performance of the sentences, such as topics extracted by topic models [22] and dependency parse trees [23]. One work which bases on pattern matching and applys extra NLP systems to derive lexical features is proposed [24], which utilizes many features derived from external corpora for a Support Vector Machine (SVM) classifier. However, all simple techniques have limitations for certain tasks. To solve the above problem, an effective solution is to factorize the linear classifier into low-rank matrices [25].

Neural network models have achieved significant success on text classification. Convolutional neural networks (CNNs) and recurrent neural networks (RNNs) have emerged as two widely used architectures and are often combined with sequence-based or tree-structured models[26,27]. CNN regards feature extraction and classification as a joint task, and extracts hierarchical representations of inputs by stacking multiple convolution kernels. Convolutional layers are similar to a sliding window over a matrix. Due to the ability of CNN to capture local features, the n-gram language model has been successfully implemented in the CNN model [28]. So far, the CNN has achieved some very successful results in the field of text classification [29, 30]. A convolution neural network architecture with multiple convolution layers is proposed, positing tent, dense and low-dimensional word vectors(initialized to random values) as inputs [31]. Experiments show that it are better than those based on unigram and bigram models. For text classification of high dimensional text, CNN achieved several state-of-the-art performances on some benchmark datasets for sentiment categorization[32]. RNN improves time complexity and analyzes texts word-by-word, considering the influence of historical sequences on current words, and can deal with the long-term dependence of a certain length of sequence, suitable for time series. Experiments show that RNN can capture long-term dependence even if there is only one single layer [33]. However, in the case where the input sequence is long, the RNN may have a gradient explosion or the gradient disappears. To avoid this problem, variants such as Long-Short-Term Memory (LSTM) [34] and Gated Recurrent Unit (GRU) [35] are designed, which have achieved some excellent results in the field of text classification [36].

Attention is an effective mechanism for selecting significant information in order to obtain superior results. Deep neural networks, including CNN and RNN, can get better result by equipped with attention mechanisms. Among many proposed attention mechanisms, some examples are excellent including soft and hard attention [37], global and local attention [38], and source-target attention and self attention [39]. In the field of image processing, by integrating pre-attention mechanisms in the optimization criterion, in the form of a saliency map, good results were obtained on the task of sequential spatial reasoning in images.

[40]. In natural language processing, attentive neural networks have achieved great success on a wide range of tasks ranging such as question answering, machine translations and tec., [41, 42, 43]. GRU compared with attention mechanisms can capture the importance of words [18]. In order to fuse the advantages of RNN, CNN and attention, the ARC model was proposed [19], and good results were obtained on the text classification task. In addition, a study has confirmed that even in the low-resource scenario, attention can be learned effectively. [44]

## III. MODEL ARCHITECTURE

In this section we will describe the details of the model framework. The model structure is shown in Fig.1. The model consists of three parts: the word embedding layer, the Pre-Attention mechanism and the post-classification net. In the word embedding layer, we get the text representation by word2vec. Then we weight the text representation with the Pre-Attention mechanism to get the attention representation. Finally, we enter the attention representation into the post-classification net for classification. In the rest of this section, we will detail the three parts above.

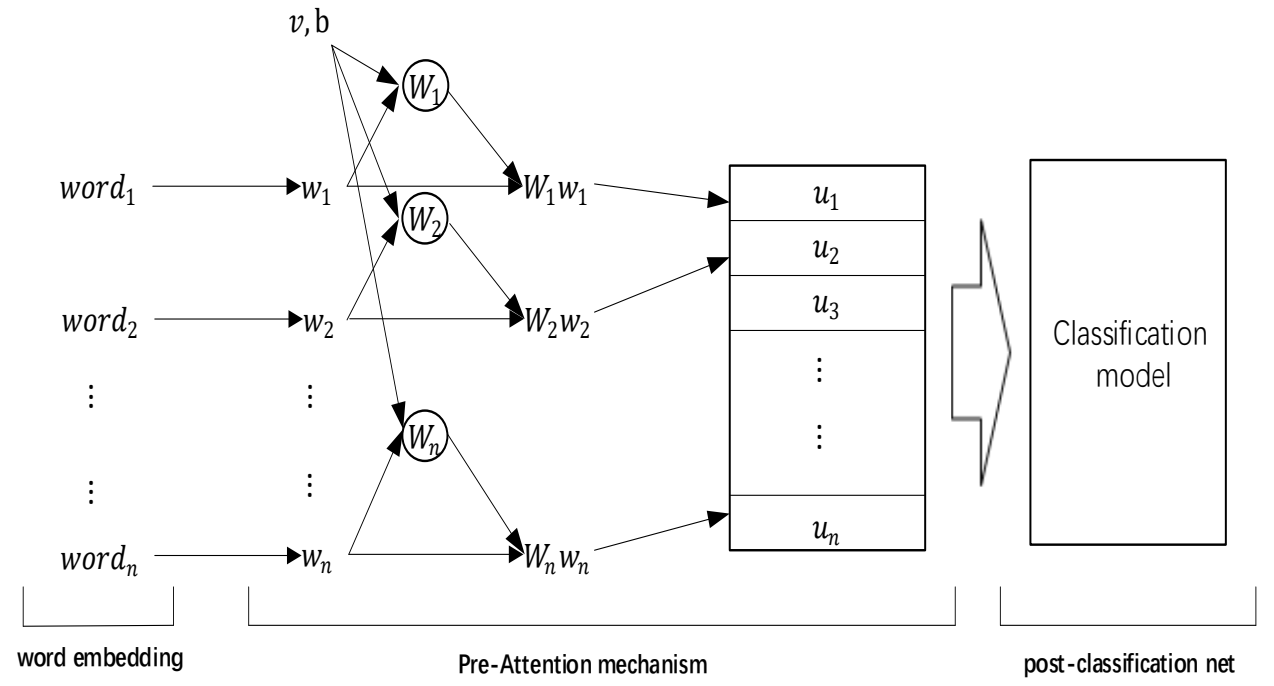

**FIGURE 1. The structure of text classification model with Pre-Attention Mechanism**

### A. THE WORD EMBEDDING LAYER

Experiments have proven that improvements in model accuracy can be obtained by performing unsupervised, pre-trained word embeddings, so we first get the word embedding matrix $M \epsilon \mathbb{R}^{|V|l}$ ($\mathbb{R}$ : The space of real numbers), where $V$ is a fixed-sized vocabulary, and $l$ is the size of word embedding. For the matrix $M$, we can initialize it with the already trained word2vec model, where $M_i$ is the vector of the word $V_i$. For the i-th word in text $word = (word_1, word_2, \dots\dots, word_n)$, we transform it into its word embedding $w_i$ by using the matrix-vector product:

$$w_i = v_i M \quad (1)$$

Where $v_i$ is a vector of size $|V|$ which has value 1 at index of $word_i$ in $V$ and 0 in all other positions. Then the sentence is feed into the next layer as a real-valued vector $S = (w_1, w_2, \dots\dots, w_n)$.

### B. PRE-ATTENTION LAYER

Not all words contribute equally to the meaning of sentences. So in this section, we impose a Pre-Attention mechanism on the inputting S to calculate the attention weight distribution of different words. This weight will make our classifiers more focused on words that play an important role in the classification task. The Pre-Attention mechanism is shown in Fig.2.

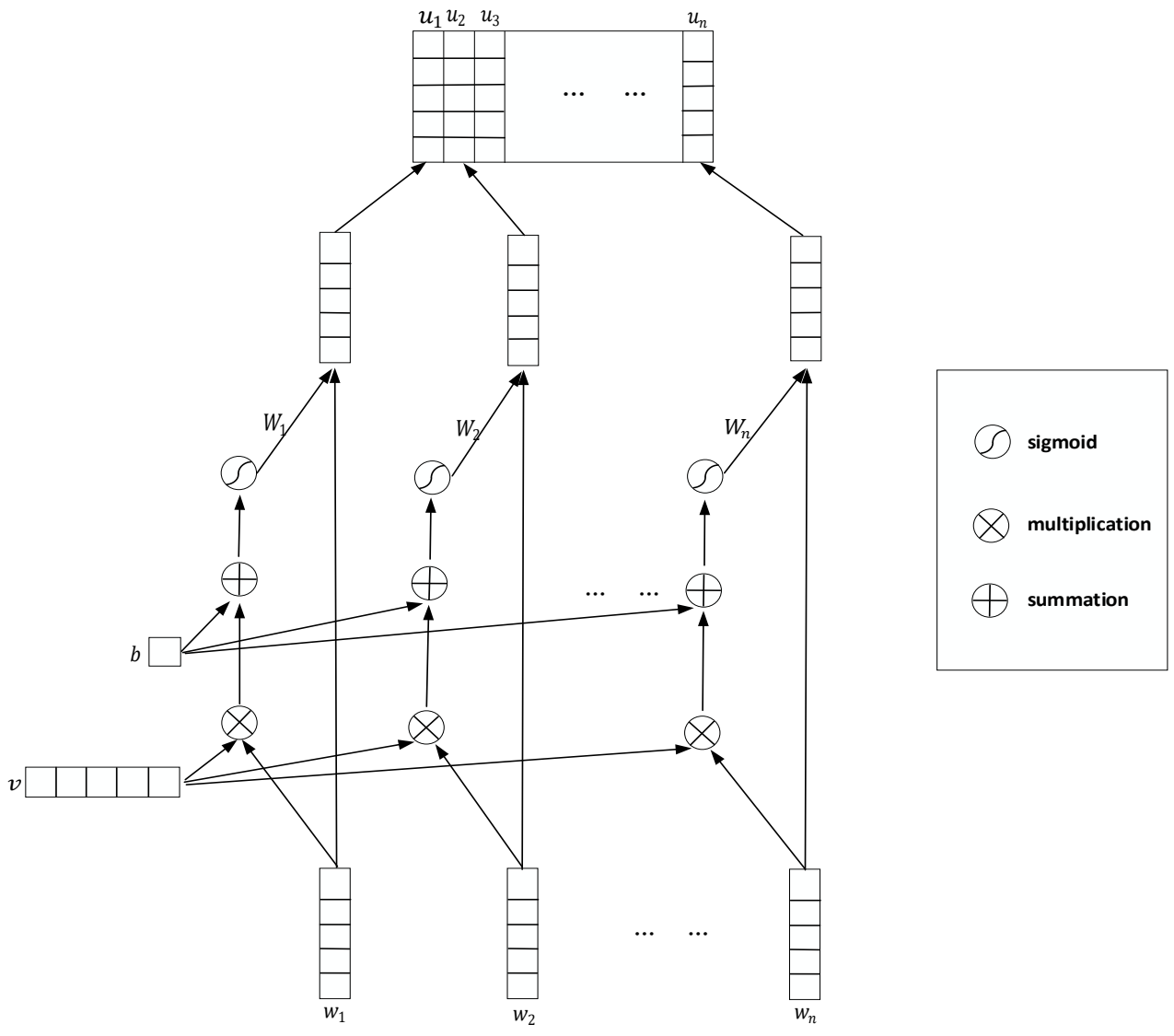

**FIGURE 2. The structure of Pre-Attention Mechanism**

For input $S$, $w_i$ represents the word vector of the i-th word, and we calculate its attention weight from:

$$W_i = f(vw_i + b) \quad (2)$$

$$f_{(x)} = \frac{1}{1 + e^{-x}} \quad (3)$$

where the attention vector $v \epsilon \mathbb{R}^l$ is a parameter to be learned. $l$ is the size of word embedding. $b \epsilon \mathbb{R}$ is a bias term to be learned and $f$ is an activation function. We utilize the $sigmoid$ as $f$. Then we use the obtained attention distribution to weight the input word vector to obtain the text representation with Pre-Attention weight, which is $U$.

$$u_i = W_i w_i \quad (4)$$

$$U = (u_1, u_2, \dots\dots, u_n) \quad (5)$$

### C. POST-CLASSIFICATION MODEL

To verify the portability of Pre-Attention mechanism, we choose two typical text classification models as post-classification networks, Text-CNN model based on CNN [28], and Att-BLSTM model based on LSTM [45].

#### 1) Text-CNN

Text-CNN is a classical model on text classification based on CNN. The model structure is shown in the Fig.3.

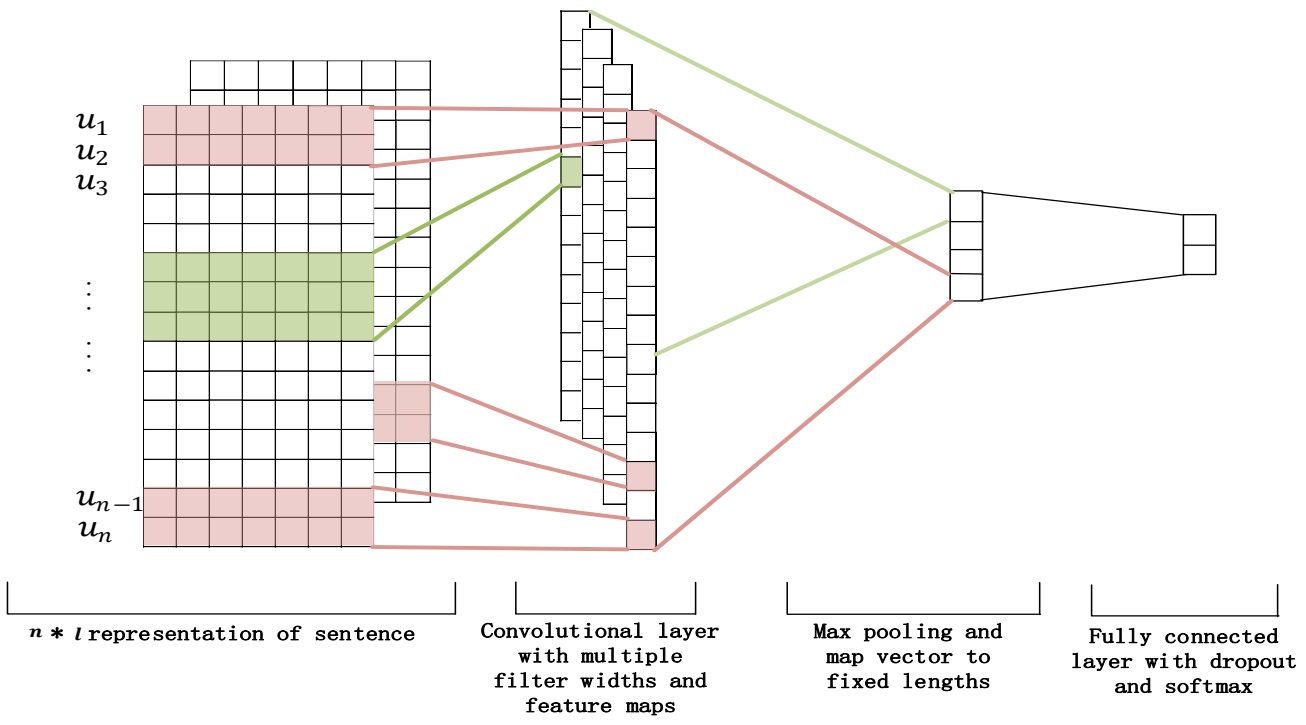

**FIGURE 3.** **The structure of Text-CNN Model**

The input is U = $(u_1, u_2, \dots\dots, u_n)$ from the Pre-Attention layer, where $u_i \epsilon \mathbb{R}^l$. $l$ is the dimension of the word vector, and $n$ is the length of the sentence. We define the following equation:

$$U_{i:j} = u_i \oplus u_{i+1} \dots\dots \oplus u_j \tag{6}$$

where $\oplus$ is a concatenation operation. A convolution operation involves a filter $c \epsilon \mathbb{R}^{hl}$ , which is applied to a window of $h$ words to produce a new feature. For example, a feature $o_i$ is generated from a window of words $U_{i:i+h-1}$ by

$$o_i = f(c \cdot U_{i:i+h-1} + b) \tag{7}$$

The convolution kernel with a height of $h$ scans input matrix U from top to bottom, equally taking n-gram feature extraction on U with size $h$. we apply a max-over time pooling operation [46] over the feature map and get maximum value of each feature map, then concatenate the maximum values into vector and feed vector to fully connection layer for classification. Meanwhile we employ dropout on the fully connection layer with a constraint on $l_2 - norms$ of the weight vectors. Finally, we get sentence category with the softmax layer.

#### 2) Att-BLSTM

Att-BLSTM is a classical model on text classification based on LSTM. The model structure is shown in the Fig.4.

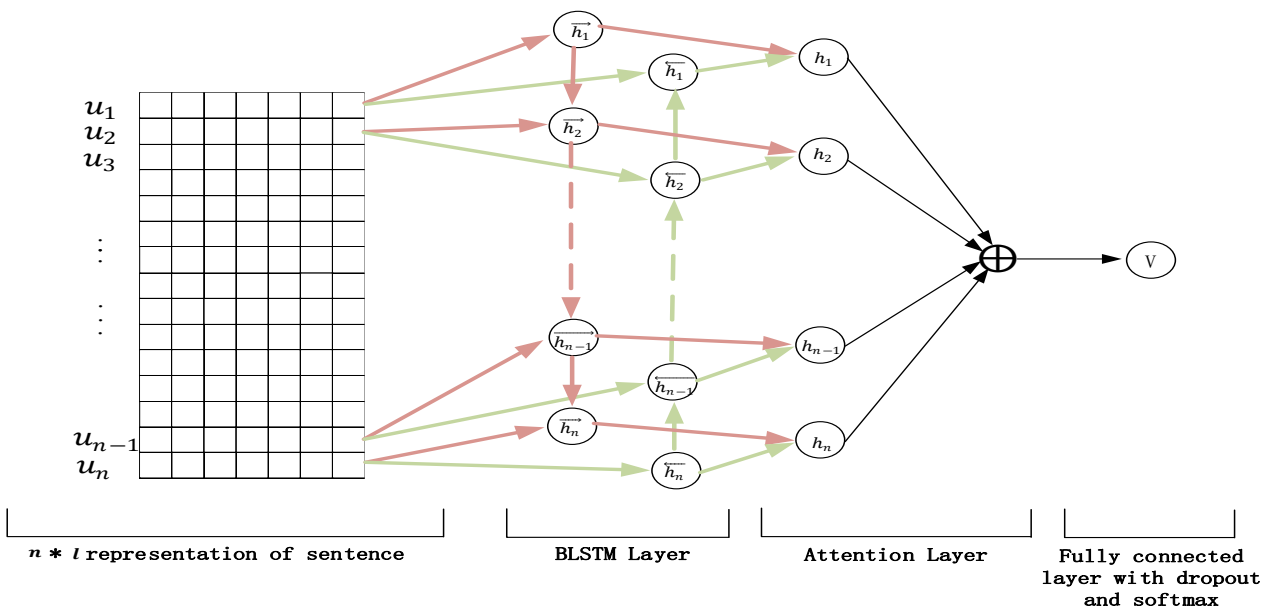

**FIGURE 4.** **The structure of Att-BLSTM Model**

Usually, as shown in Fig.5, the LSTM unit contains three parts.

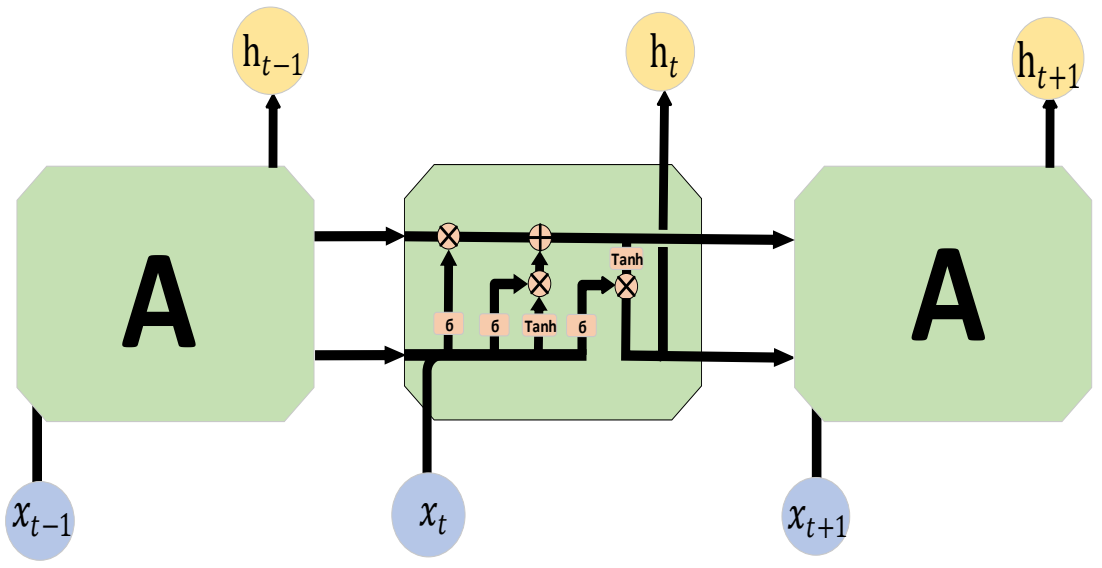

**FIGURE 5.** **Unit of Long Short-Term Memory**

One forget gate $f_t$ with corresponding weight matrix $W_f$, $b_f$, which decides which state information to discard:

$$f_t = \sigma\left(W_f[h_{t-1}, x_t] + b_f\right) \tag{8}$$

One input gate $i_t$ and $\widetilde{C}_t$ with corresponding weight matrix $W_i, W_C, b_i, b_C$, which decides which kind of cell state should be added:

$$i_t = \sigma(W_i[h_{t-1}, x_t] + b_i) \tag{9}$$
$$\widetilde{C}_t = \tanh(W_C[h_{t-1}, x_t] + b_C) \tag{10}$$

Then, cell states can be updated by follow equation. Where $f_t * C_t$ forgets the state information what we want to discard, and $i_t * \widetilde{C}_t$ adds new content we want to remember.

$$C_t = f_t * C_{t-1} + i_t * \widetilde{C}_t \tag{11}$$

One output gate $O_t$ with corresponding weight matrix $W_o$, $b_o$, which decides which part of the cell state to output.

$$O_t = \sigma(W_o[h_{t-1}, x_t] + b_o) \tag{12}$$
$$h_t = O_t * \tanh(C_t) \tag{13}$$

For many sequence modelling tasks, it is beneficial to have access to future as well as past context. However, standard LSTM networks ignore future context, and they process sequences in temporal order. To address this limitation, Bidirectional LSTM networks extend the unidirectional LSTM networks by introducing a second layer, where the hidden to hidden connections flow in opposite temporal order. Therefore, the model is able to exploit information both from the future and the past.

As shown in Fig.4, We first enter $U$ into a two-way lstm network, which contains two sub-networks for the left and right sequence context. Then we can get the output of the i-th word, which is shown in the following equation:

$$h_i = \overrightarrow{h_i} \oplus \overleftarrow{h_i} \tag{14}$$
$$H = [h_1, h_2 \dots\dots h_n] \tag{15}$$

Then we pay attention mechanism to the output vectors $H$ that the BLSTM layer produced. where $w \epsilon R^l$ is a trained parameter vector, and $l$ is the size of word embedding.

$$\mathrm{M} = \tanh(\mathrm{H}) \quad (16)$$

$$\alpha = \mathrm{softmax}(w^T M) \quad (17)$$

Then we get the sentence representation γ by a weighted sum of those output vectors:

$$\gamma = \mathrm{H}\alpha^T \quad (18)$$

Finally, we obtain the final sentence representation by the following equation:

$$h^* = \tanh(\gamma) \quad (19)$$

Then we feed vector $h^*$ to fully connection layer for classification. Meanwhile we employ dropout on the fully connection layer with a constraint on $l_2 - norms$ of the weight vectors. Finally, we get sentence category with the softmax layer.

## IV. EXPERIMENTAL SETUP

### *A. DATASETS*

We test the network on three different datasets, whose details are shown in Table 1.

- **Stanford Large Movie Review dataset (IMDB):** The IMDB [47] consists of 50,000 binary labeled reviews, which are divided into 50:50 training and testing sets. The distribution of labels is balanced in each sub-dataset. One key aspect of this dataset is that there are several sentences in each review.
- **Subjectivity dataset (Subj):** Subjectivity dataset [48] where the task is to classify a sentence as being subjective or objective, which is collected from snippets of movie reviews from Rotten Tomatoes and plot summaries for movies from the Internet Movie Database. It consists of 10000 binary labeled reviews, including 5000 subjective reviews and 5000 objective reviews.
- **Movie reviews (MR):** The database [49] consists of 10662 reviews from Rotten Tomatoes webpages, including 5,331 positive and 5,331 negative samples. Every review includes one sentence. Those reviews marked with "fresh" are positive, and those reviews marked with "rotten" are negative

For the IMDB, we used 20% of the labeled training documents as a validation set. For the Subj, we split it into three sets: 7k sentences for training, 2k sentences for testing, and 1k sentences as a validation set. The MR is also splited into three sets: 70% of the sentences for training, 20% of the sentences for testing, and 10% of the sentences being validation set. In Table 1, we present additional details about the three benchmark datasets.

### *B. HYPERPARAMETER AND TRAINING*

#### 1) MODEL VARIATIONS

We experiment with several variants of the model

- **Pre-Attention-Text-CNN:** A model uses Text-CNN as a post-classification model with Pre-Attention, using pre-trained vectors from word2vec in the word embedding layer. For those unknown words, we randomly initialize them. The word embedding matrix $\boldsymbol{M}$ are kept static and only the other parameters can be learned.
- **Text-CNN:** Same as above but the Pre-Attention mechanism is removed.
- **Pre-Attention-Att-BLSTM:** A model uses Att-BLSTM as a post-classification model with Pre-Attention, using pre-trained vectors from word2vec in the word embedding layer. For those unknown words, we randomly initialize them. The word embedding matrix $\boldsymbol{M}$ are kept static and only the other parameters can be learned.
- **Att-BLSTM:** Same as above but the Pre-Attention mechanism is removed.

#### 2) THE WORD EMBEDDINGS

In our experiments, we utilized the publicly available word2vec vectors that were trained on 100 billion words from Google News. The vectors were trained using the continuous bag-of-words architecture [50]. The size of word enbedding are optional, and we use vectors with dimensional of 300.

#### 3) Text-CNN

For all datasets we use: windows ($h$) of 1,2,3,4,5 with 128 feature maps each, dropout rate of 0.4, the L2 regularization of 1, the learning rate of 0.001, the mini-batch of 64. All of above values were chosen via a grid search on the IMDB with Text-CNN.

#### 4) Pre-Attention-Text-CNN

Except for Pre-Attention mechanism, the parameters of the other part are the same as above.

#### 5) Att-BLSTM

For all datasets we use: dropout rate of 0.5, the L2 regularization of 0.1, the learning rate of 0.01, the mini-batch of 64, the hidden layer size of LSTM of 50. All of above values were chosen via a grid search on the IMDB with Att-BLSTM

#### 6) Pre-Attention-Att-BLSTM

Except for Pre-Attention mechanism, the parameters of the other part are the same as above.

## V. EMPIRICAL RESULTS AND ANALYSIS

### *A. MODEL AND RESULTS*

Table 2 shows the results of our models and other state-of-the-art methods of text classification.

The optimal model on the IMDB dataset is Pre-Attention-Text-CNN, and its classification accuracy is only 0.2% lower than DSCNN-Pretrain. Compared with other models, the classification accuracy has increased by 0.3%-7.31%. For the

**TABLE 1. Summary statistics for the datasets. $T$: The type of review, $c$: Number of target classes. $L$: Average sentence length. $N$: Dataset size. $|V|$: Vocabulary size. $Train$: Train set size. $Test$: Test set size. $Dev$: Validation set size.**

| ***Data*** | $T$ | $c$ | $L$ | $N$ | $\|V\|$ | $Train$ | $Test$ | $Dev$ |
|---|---|---|---|---|---|---|---|---|
| IMDB | Document | 2 | 230 | 50000 | 89527 | 20k | 25k | 5k |
| Subj | Sentence | 2 | 23 | 10000 | 21323 | 7k | 2k | 1k |
| MR | Sentence | 2 | 20 | 10662 | 18765 | 7464 | 2132 | 1066 |

**TABLE 2. The classification accuracy (%) of our model compared to other approaches on IMDB, MR and Subj.**

| *Method* | *IMDB* | *MR* | *Subj* |
|---|---|---|---|
| Text-CNN | 89.0 | 79.9 | 92.5 |
| Pre-Attention-Text-CNN | **90.5** | **82.3** | **93.7** |
| Att-BLSTM | 86.5 | 79.1 | 90.3 |
| Pre-Attention-Att-BLSTM | 88.3 | 80.0 | 92.1 |
| DNN[51] | 88.55 | - | - |
| Naïve bayes classifier[51] | 83.19 | - | - |
| RNN[52] | 88.59 | - | - |
| CNN[52] | 87.44 | - | - |
| Svm[53] | 87.97 | - | - |
| SVM(TF-IDF)[54] | 88.45 | - | - |
| DSCNN [55] | 90.2 | 81.5 | 93.2 |
| DSCNN-Pretrain[55] | **90.7** | **82.2** | **93.9** |
| ESN [56] | - | 78.1 | 92.6 |
| CNN-BiGRU[57] | - | 79.4 | 93.8 |
| CNN-Ana[58] | - | 81.02 | 93.66 |
| DSCNN[59] | - | 81.5 | - |
| combine-skip[60] | - | - | 93.6 |

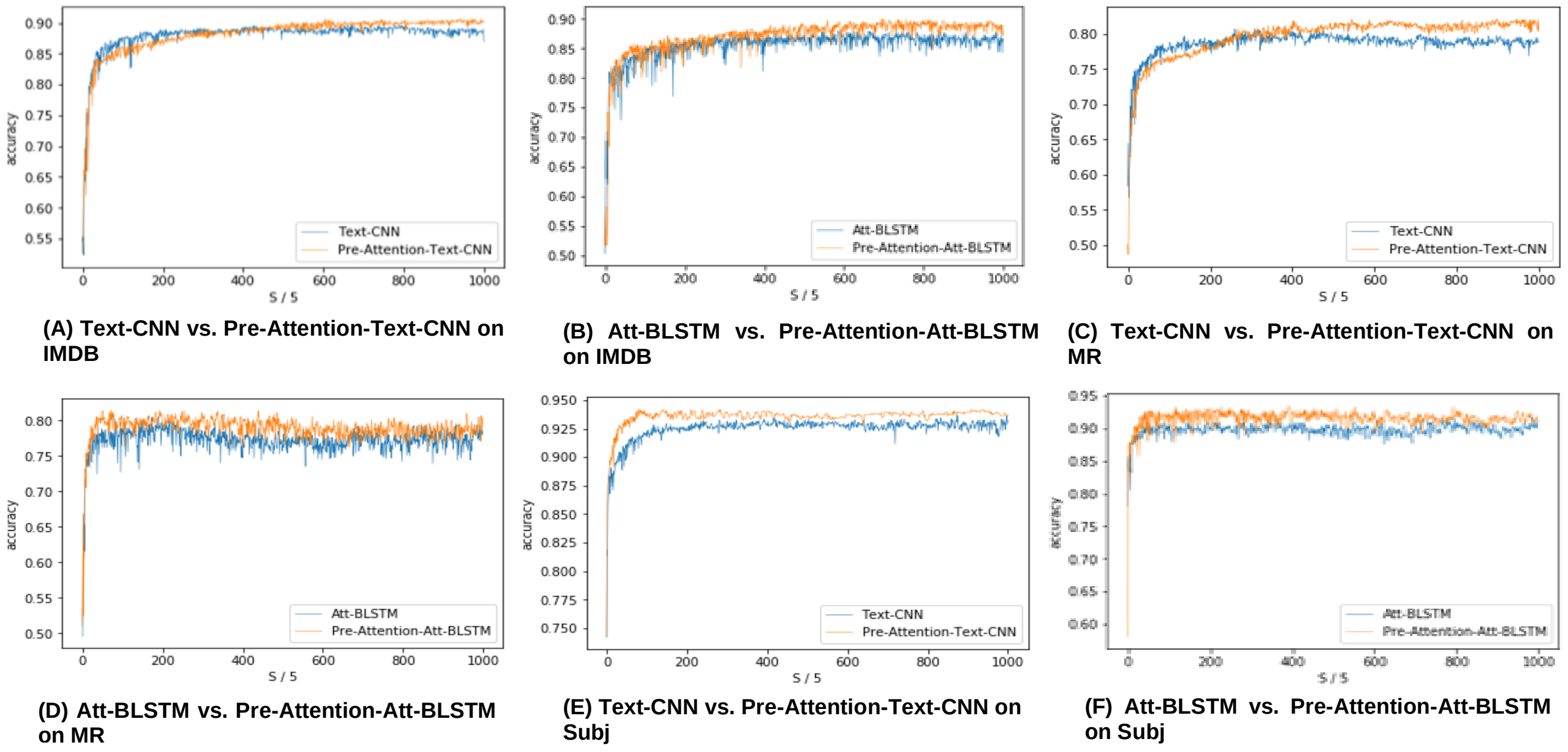

**(A) Text-CNN vs. Pre-Attention-Text-CNN on IMDB**

**(B) Att-BLSTM vs. Pre-Attention-Att-BLSTM on IMDB**

**(C) Text-CNN vs. Pre-Attention-Text-CNN on MR**

**(D) Att-BLSTM vs. Pre-Attention-Att-BLSTM on MR**

**(E) Text-CNN vs. Pre-Attention-Text-CNN on Subj**

**(F) Att-BLSTM vs. Pre-Attention-Att-BLSTM on Subj**

**FIGURE 6. Results of comparing Pre-Attention-Classification model and Single-Classification model on three datasets (IMDB, MR, Subj),**

Movie reviews (MR) dataset, our optimal model is Pre-Attention-Text-CNN, and the classification accuracy reaches 82.3%, compared with several other classification models on MR. Accuracy has increased by 0.1%-4.2%. For the subj dataset, each review containing a sentence. The Pre-Attention-Text-CNN is still the best performer, and the accuracy rate is 93.7%. Compared with several other comparison models, the model accuracy rate exceeds the model other than DSCNN-Pretrain. The experimental results verified the effectiveness of Pre-Attention mechanism.

Although it is generally believed that RNN is more suitable for NLP tasks, it can better refer to word order information. But in this paper, we find that the CNN-based model is better than the RNN-based model. This may be because for the classification task of this paper, the phrase with significant emotional polarity will have a more critical impact on the result. CNN mainly does the extraction of local features, similar to n-gram, so it is understandable that the CNN network can work better in the tasks of this paper.

### B. PRE-ATTENTION-CLASSIFICATION MODEL VS. SINGLE-CLASSIFICATION MODEL

In order to prove the effectiveness of the Pre-Attention mechanism, we compared the classification accuracy rate from Pre-Attention-classification model and single-classification model on the same dataset. So there are two comparative experiments, Text-CNN vs. Pre-Attention-Text-CNN and Att-BLSTM vs. Pre-Attention- Att-BLSTM. We performed our experiments on the above three datasets. For every comparison, we train the two classifier models and calculate the classification accuracy on the test set every 5 steps. The experimental results are shown in the Fig.6, where $S$ is the number of training steps. X-axis represents $\frac{S}{5}$, and the Y-axis represents the classification accuracy on the test set. We can see that the Pre-Attention mechanism improves the accuracy of the classification model. According to Table 2, we find that compared with the text classification model without Pre-Attention mechanism, those with Pre-Attention mechanism improved accuracy by 0.9%-2.4%, which clearly demonstrates the effectiveness of the proposed Pre-Attention mechanism.

### C. VISUALIZATION OF PRE-ATTENTION

Another advantage of Pre-Attention is that it is easier to visualize, which is very instructive for us to analyze which words are more important for classification. We randomly select some texts from the above three datasets, calculating the pre-attention value, and visualize the word attention $W_t$ using an open source sequence annotation tool [61], visualized in Fig.7. We can find that Pre-Attention models give more attention to words with strong emotions and degree adverbs, such as *absolutely, horrible, sure, like, serious, good, better, successful*, etc., which proves that Pre-Attention mechanism can learn explicit emotional tendencies in sentences and successfully integrates an emotional lexicon into deep neural network. The Pre-Attention mechanism can learn explicit emotional tendencies in sentences and have a good visualization.

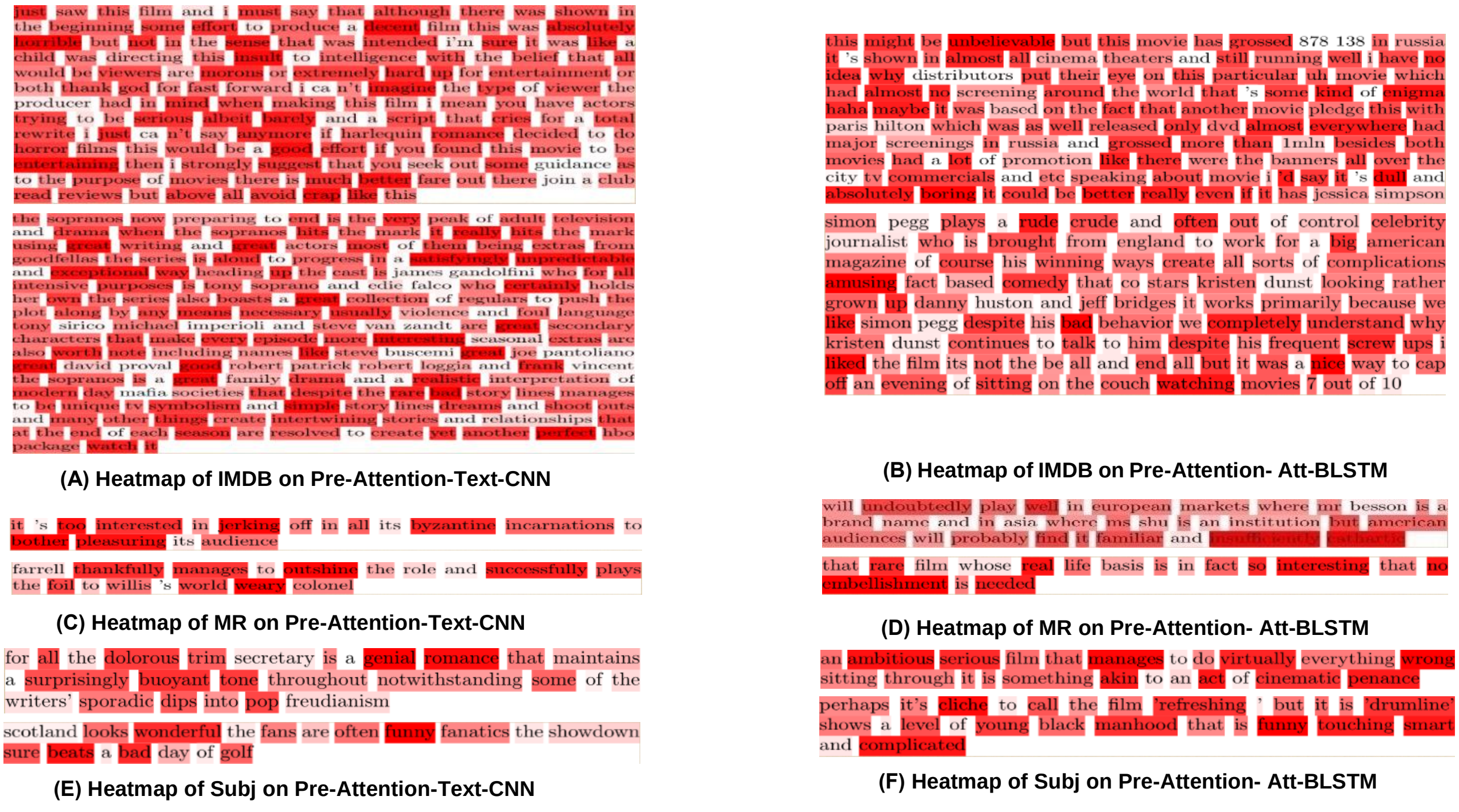

(A) Heatmap of IMDB on Pre-Attention-Text-CNN

(B) Heatmap of IMDB on Pre-Attention- Att-BLSTM

(C) Heatmap of MR on Pre-Attention-Text-CNN

(D) Heatmap of MR on Pre-Attention- Att-BLSTM

(E) Heatmap of Subj on Pre-Attention-Text-CNN

(F) Heatmap of Subj on Pre-Attention- Att-BLSTM

**FIGURE 7. Heatmap of three datasets (IMDB, MR, Subj) on Pre-Attention-Classification model (Pre-Attention-Text-CNN, Pre-Attention-Att-BLSTM)**

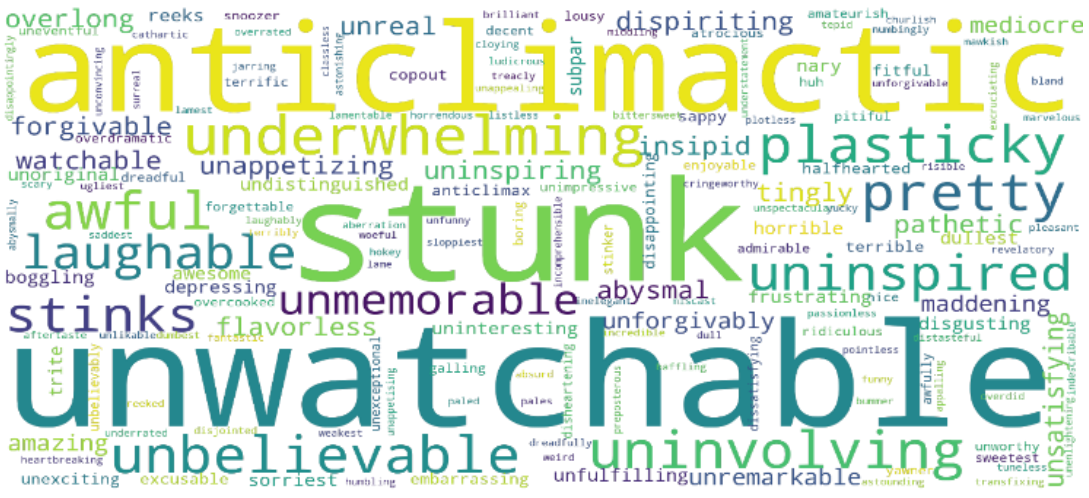

(A) Word clouds of IMDB on Pre-Attention-Text-CNN

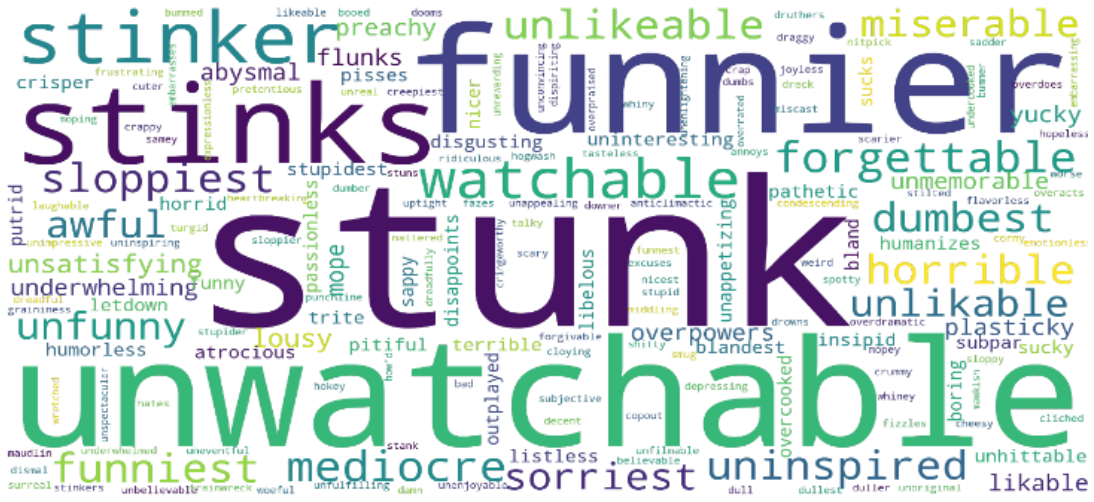

(B) Word clouds of IMDB on Pre-Attention-Att-BLSTM

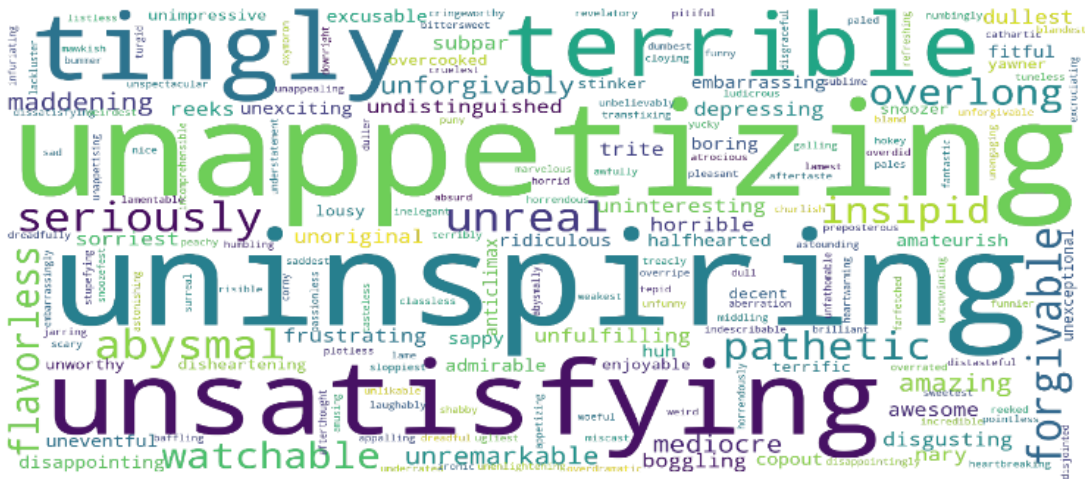

(C) Word clouds of MR on Pre-Attention-Text-CNN

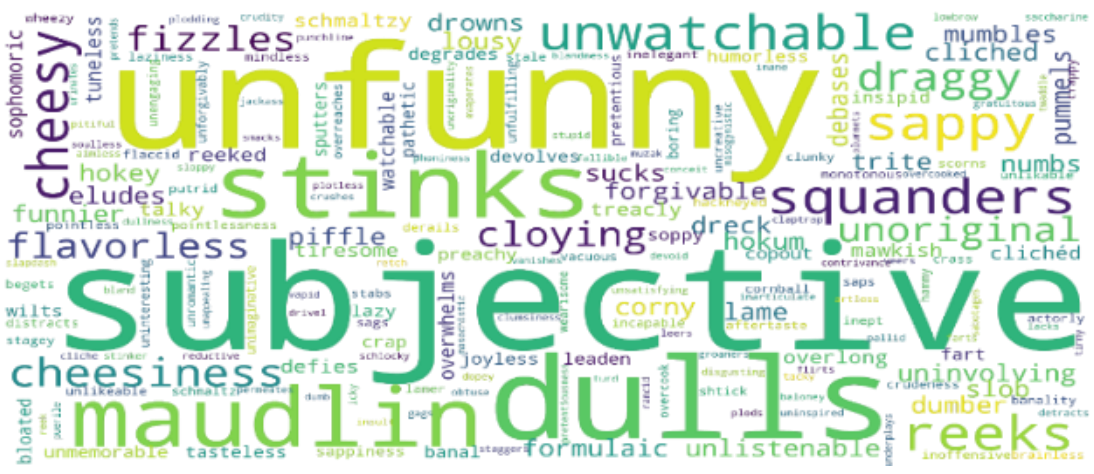

(D) Word clouds of MR on Pre-Attention-Att-BLSTM

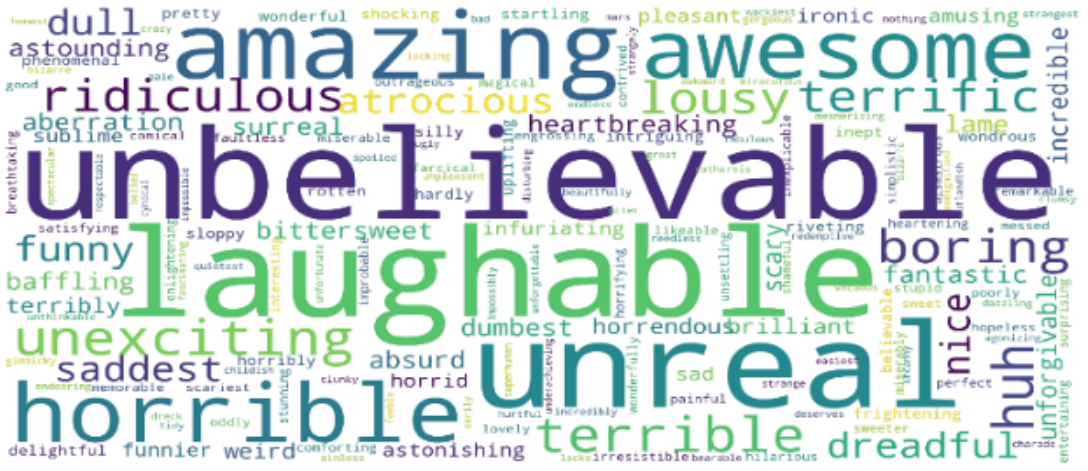

(E) Word clouds of Subj on Pre-Attention-Text-CNN

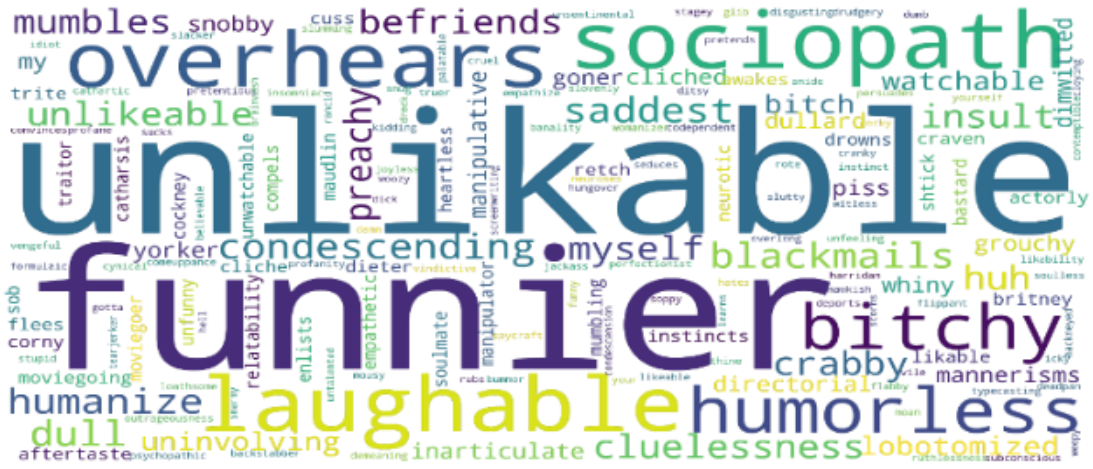

(F) Word clouds of Subj on Pre-Attention-Att-BLSTM

**FIGURE 8. Word clouds of three datasets (IMDB, MR, Subj) on Pre-Attention-Classification model (Pre-Attention-Text-CNN, Pre-Attention-Att-BLSTM)**

### *D. SENTIMENT LEXICON*

For above classification tasks, after getting the words' attention values by Pre-Attention mechanism, we draw word clouds based on attention values, which are shown in Fig.8. We can find those words with strong emotions and degree adverbs get more attention, such as *funnier*, *anticlimactic*, *terrible*, *unappetizing*, *unsatisfying*, *dull*, *amazing*, etc., which proves that Pre-Attention mechanism can learn explicit emotional tendencies in sentences. Then we prove that we have got an excellent lexicon from two aspects of stability and effectiveness.

#### *1) EFFECTIVENESS*

In order to prove the validity of the lexicon extracted by the Pre-Attention value, we compare it with a handcrafted lexicon. Subjectivity Lexicon [62] is a lexicon including 8222 words. Every word has two type of labels. In Table 3, we present additional details about the dataset.

We select the words belonging to strongsubj to form subj-lexicon set $S_l$ and those words being positive, negative or both priorpolaritys to form priorpolaritys-lexicon set $P_l$. Then we define $W^L = (W_1^L, W_2^L \dots\dots, W_k^L)$ to represent sequences of words sorted by Pre-Attention values, where the Pre-Attention value of $W_i^L$ is bigger than the Pre-Attention value of $W_j^L$ when $i$ is smaller than $j$. We define the equation (21) to measure validity of lexicon by Pre-Attention mechanism.

**TABLE 3. Summary statistics for the Subjectivity Lexicon. $w$: The number of weaksubj words, $s$: The number of strongsubj words. $p$: The number of positive words. $n$: The number of negative words. $b$: The number of words with both priorpolaritys. $N$: The number of neutral words.**

| | *type* | | *priorpolarity* | | | |
|---|---|---|---|---|---|---|
| | $w$ | $s$ | $p$ | $n$ | $b$ | $N$ |
| **Subjectivity Lexicon** | 2653 | 5569 | 2718 | 4912 | 21 | 571 |

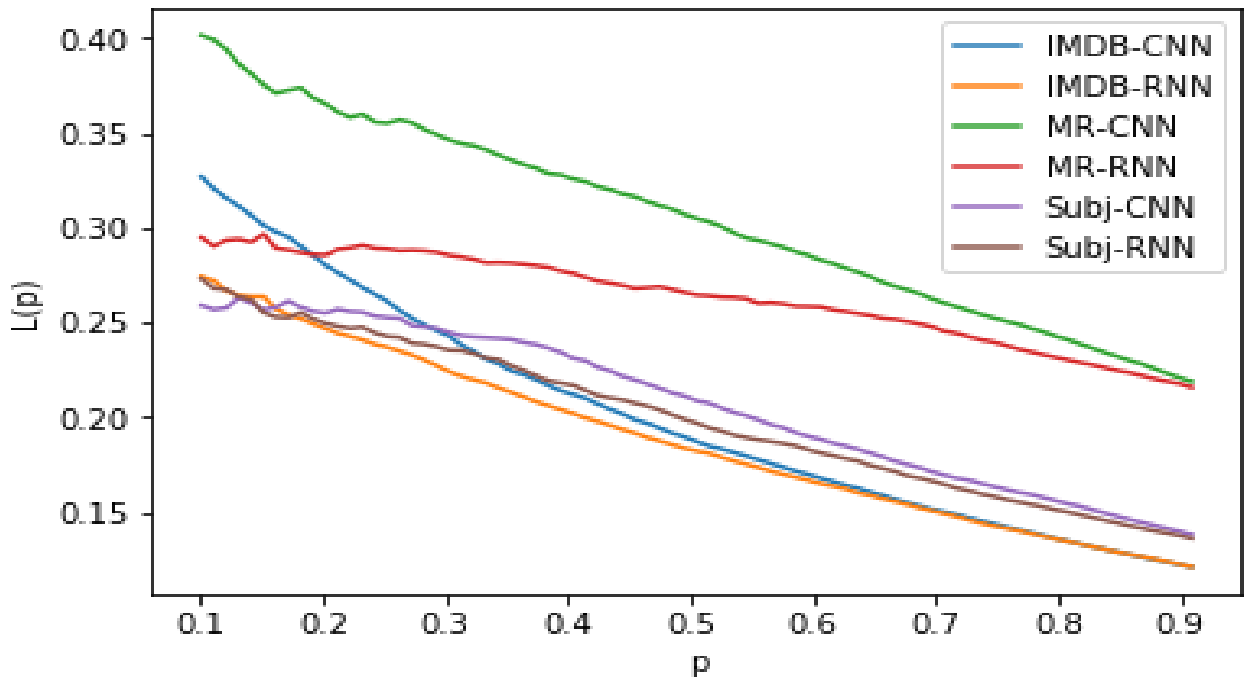

**FIGURE 9. The similarity of lexicons from Pre-Attention mechanism and handcrafted lexicons with threshold change.**

$$D_p = \left(W_1^L, W_2^L \ldots \ldots, W_{\lfloor k*p \rfloor}^L\right)(0 < p < 1) \quad (20)$$

$$L_{(p)} = \frac{|(D_p \cap D)|}{\lfloor k*p \rfloor} \quad (21)$$

$L_{(p)}$ represents the ratio of words belonging to $D$ in $D_p$. Where $D$ is a handcrafted lexicon. When $W^L$ is from IMDB or MR, we set $D$ to $P_l$. When $W^L$ is from Subj, we set $D$ to $S_l$. Fig.9 shows the change of $L_{(p)}$ with $p$ in different situations.

Since the handcrafted lexicon is a general lexicon rather than a lexicon like the one we extracted by Pre-Attention mechanism, which is for a data set, $L_{(p)}$ is not high. But we can still prove from the trend of $L_{(p)}$ that those words with higher attention values are more likely to appear in the lexicon for classification, which proves that the Pre-Attention value can reflect the explicit emotional tendencies in the sentences.

### *2) STABILITY*

For a text classification task, we believe that each word in the dataset has a different impact on the classification results, and there is a unique ordering that indicates the importance of the words for this task. Although the importance of a word for particular two sentences may be different, in a statistical sense, our assumptions are reasonable for the entire classification task.

Next, in order to prove the stability of the pre-attention mechanism, we compare the ordering of words sorted by the Pre-Attention weight from the two pre-attention models (Pre-Attention-Text-CNN and Pre-Attention-Att-BLSTM) in a dataset classification task.

We define $o_c$ and $o_r$ to represent sequences of words index sorted by Pre-Attention values, where $o_c$ and $o_r$ are from Pre-Attention-Text-CNN and Pre-Attention-Att-BLSTM respectively, and $k$ is the number of words of the dataset.

$$o_c = (o_c^1, o_c^2 \ldots \ldots o_c^k) \quad (22)$$

$$o_r = (o_r^1, o_r^2 \ldots \ldots o_r^k) \quad (23)$$

By equation (2), we get:

$$W_{o_m^i} < W_{o_m^j} \ (m \in \{c, r\}, i < j) \quad (24)$$

Then we define a threshold to get sentiment lexicons. we put the words of the top $(1-p)(0 < p < 1)$ of the Pre-Attention value into a lexicon. We get:

$$D_c = \left(o_c^{\lfloor k*p \rfloor}, o_c^{\lfloor k*p \rfloor + 1} \ldots \ldots o_c^k\right) \quad (25)$$

$$D_r = \left(o_r^{\lfloor k*p \rfloor}, o_r^{\lfloor k*p \rfloor + 1} \ldots \ldots o_r^k\right) \quad (26)$$

$D_c$ is a sentiment lexicon from Pre-Attention-Text-CNN, and $D_r$ is a sentiment lexicon from Pre-Attention-Att-BLSTM. We define the following equation (27) to measure the similarity of the sentiment lexicons.

$$y_{(p)} = \frac{|(D_c \cap D_r)|}{\lceil k*(1-p) \rceil} \quad (27)$$

Fig.10 depicts the similarity of sentiment lexicons $D_c$ and $D_r$ on three datasets. We also added the similarity of sentiment lexicons when $D_c$ and $D_r$ are two random sequences, where every element is between 1 and $k$, and there are no two same numbers in every sequence. In the figure, the x-axis represents $p$ and y-axis is $y_{(p)}$. This figure shows that the sentiment lexicon extracted by our Pre-Attention model has good stability even with different post-classifiers.

Then we define equation (28) to measure the relative stability of the sentiment lexicons, where $(1-p)$ is the lexicons similarity, when $D_c$ and $D_r$ are two random sequences. We describe the change of $Y_{(p)}$ with $p$ in Fig.11. We can see that $Y_{(p)}$ continues to grow with $p$ growing, which means that those words with high Pre-Attention values are more easily identified by the Pre-Attention mechanism. In Table 4, we list the values of $y_{(p)}$ and $Y_{(p)}$ when $p$ is 0.5, 0.6, 0.7, 0.8, 0.9 on three datasets.

$$Y_{(p)} = \frac{y_{(p)}}{(1-p)} \quad (28)$$

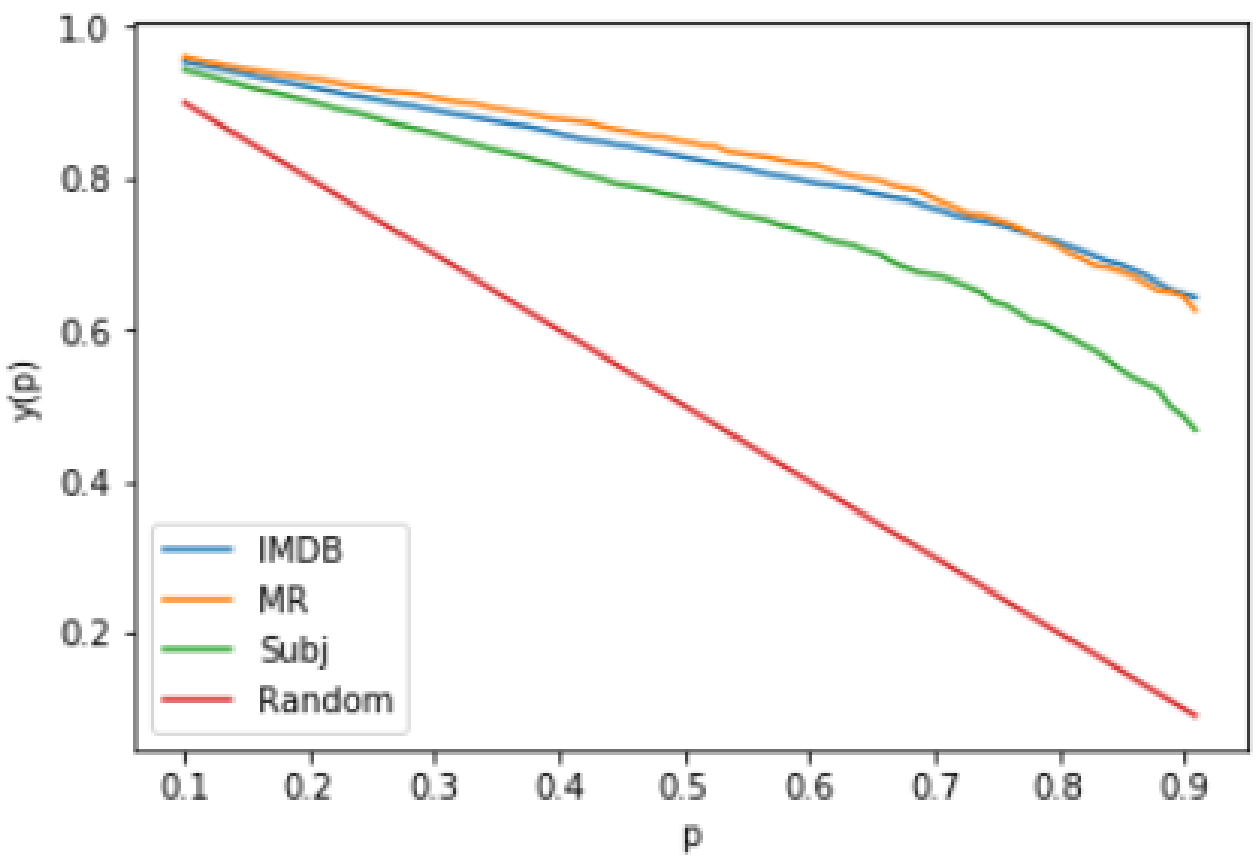

**FIGURE 10. The similarity of sentiment lexicons from Pre-Attention-Text-CNN and Pre-Attention-Att-BLSTM on three datasets with threshold change.**

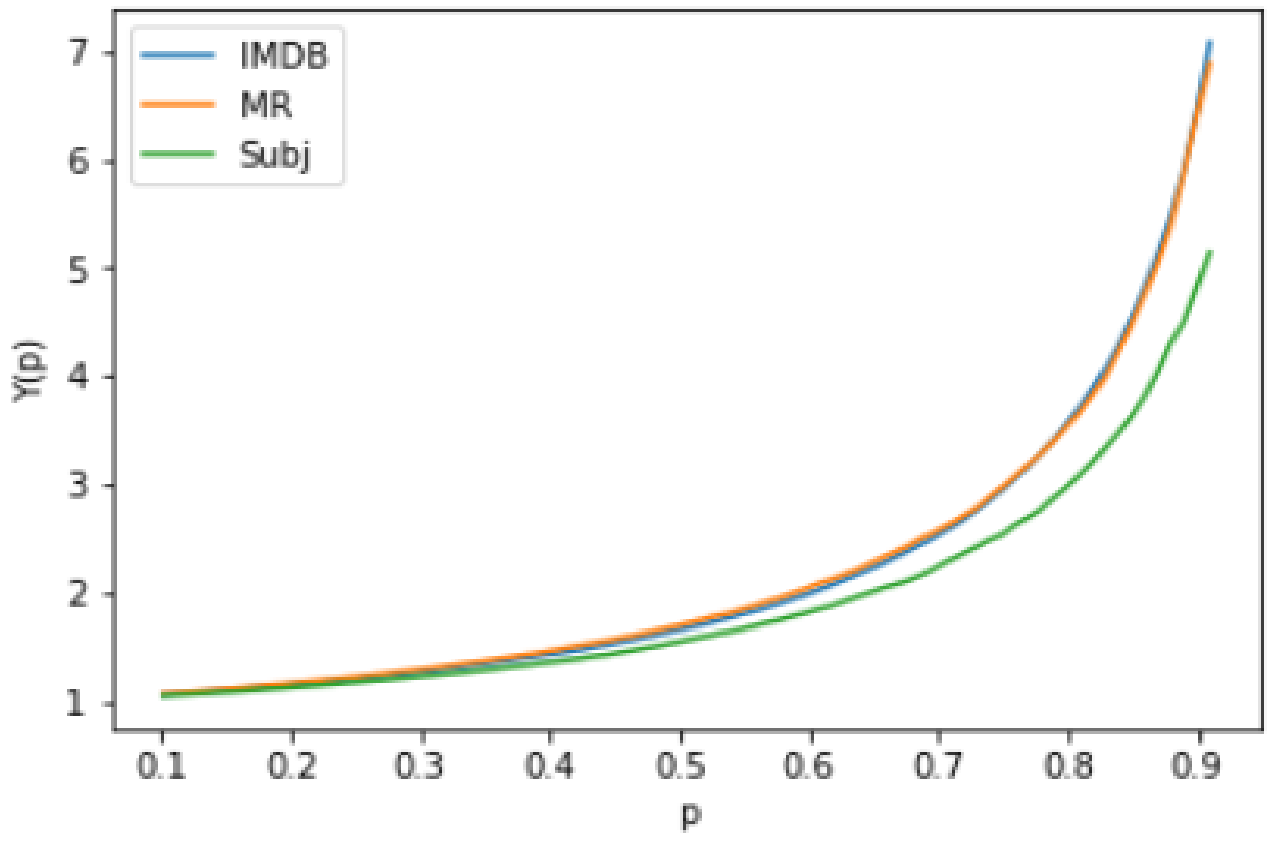

**FIGURE 11. The relative similarity of sentiment lexicons from Pre-Attention-Text-CNN and Pre-Attention-Att-BLSTM on three datasets with threshold change.**

**TABLE 4. The values of $y_{(p)}$ and $Y_{(p)}$ when $p$ is 0.5, 0.6, 0.7, 0.8, 0.9 on three datasets**

| | *IMDB* | | *MR* | | *Subj* | |
|---|---|---|---|---|---|---|
| $p$ | $y_{(p)}$ | $Y_{(p)}$ | $y_{(p)}$ | $Y_{(p)}$ | $y_{(p)}$ | $Y_{(p)}$ |
| 0.5 | 82.61% | 1.669 | 84.86% | 1.712 | 82.24% | 1.559 |
| 0.6 | 79.30% | 2.012 | 81.68% | 2.073 | 78.54% | 1.839 |
| 0.7 | 75.66% | 2.582 | 76.73% | 2.619 | 74.03% | 2.286 |
| 0.8 | 71.02% | 3.700 | 70.11% | 3.653 | 67.12% | 3.076 |
| 0.9 | 64.34% | 7.077 | 62.63% | 6.888 | 55.86% | 5.144 |

In summary, the Pre-Attention mechanism assigns those words with strong emotions and degree adverbs more attention,

even with different post-classification models, which proves the Pre-Attention mechanism has great effectiveness, stability and portability. By setting proper thresholds, we can obtain reliable and stable sentiment lexicons.

## VI. CONCLUSION

For the problem that the linguistic knowledge is not fully utilized in the text classification task, in this paper, we presented the Pre-Attention mechanism. It can automatically assign different attention values to words according that different importance for the text classification task, equally integrating a lexicon for classification into deep neural networks. Our approach performed well on three benchmark datasets. Our results demonstrated that those models with Pre-Attention mechanism achieved higher accuracy on all three datasets compared with those models without Pre-Attention, which proves the validity of Pre-Attention mechanism. When comparing with other several methods, our approach also performed well and achieved a competitive classification accuracy.

In addition, those words with high Pre-Attention values are more likely to be in lexicons for classification, and we got lexicons by those attention values, which is an inspiring method of information extraction. And we proved the stability of the lexicons extracted by Pre-Attention mechanism. However, there are still some limitations on the Pre-Attention mechanism. For example, we can only get the attention value of a single word. In the future, we plan to extend Pre-Attention mechanism to 2-gram, 3-gram and etc.